\documentclass{article}

\PassOptionsToPackage{numbers,compress}{natbib}

\usepackage[preprint]{neurips_2026}
\usepackage{tabularx} 

\usepackage[utf8]{inputenc} 
\usepackage[T1]{fontenc}    
\usepackage{hyperref}       
\usepackage{url}            
\usepackage{booktabs}       
\usepackage{amsfonts}       
\usepackage{nicefrac}       
\usepackage{microtype}      
\usepackage{xcolor}         
\usepackage{graphicx}

\title{EmoMind: Decoding Affective Captions from Human Brain fMRI}

\author{%
  Bilal A. Mohammed \\
  Department of Biomedical Engineering\\
  Vanderbilt University\\
  Nashville, TN \\
  \texttt{bilal.mohammed@vanderbilt.edu} \\
  \And
  Lin Gu \\
  Research Institute of Electrical Communication\\
  Tohoku University\\
  Sendai, Japan \\
  \texttt{lin.gu@riken.jp} \\
  \And
  Ruogu Fang \\
  Department of Biomedical Engineering\\
  University of Florida\\
  Gainesville, FL \\
  \texttt{Ruogu.Fang@bme.ufl.edu} \\
}

\begin{document}

\maketitle

\begin{abstract}
Decoding visual experience from brain activity has advanced substantially,
but current brain-to-text systems largely recover semantic content while
discarding affect. Additionally, language models can generate emotional
text when prompted with categorical labels, but such labels collapse rich
inter-subject variability into coarse discrete bins. We present EmoMind,
the first end-to-end pipeline for decoding affective captions directly
from fMRI signals. EmoMind first retrieves a semantically grounded neutral
scene description from brain-decoded visual features, then rewrites it
using a continuous 34-dimensional emotion vector decoded from the same
fMRI recording. To control the balance between content preservation and
affective expression, we train the rewriter with classifier-free guidance
against an identity-preserving null branch, enabling smooth interpolation
between semantic fidelity and affective expressivity. We evaluate affective caption generation with a three-axis validation
framework spanning subject-specificity, structural geometry, and causal
control. We further augment this framework with a synthetic-brain
substitution test that probes robustness to the measurement apparatus, and
we benchmark each axis against GPT-4 prompted with brain-decoded top-5
emotion labels as a strong discrete baseline. Across two independent
emotion fMRI datasets, EmoMind significantly outperforms label-prompted
GPT-4 on all three axes, with the largest gains on metrics that require
person-specific affective structure rather than population-level emotion
aggregation. These results establish continuous brain-decoded affect as a
viable control signal for individualized affective caption generation and
open new directions for studying individual affective brain organisation.
\end{abstract}

\section{Introduction}

Advances in non-invasive neuroimaging techniques, like functional magnetic resonance imaging (fMRI), have allowed researchers to measure patterns of brain
activity linked to high-level cognitive functions such as language processing and emotion perception. Converting neural signals into emotion expressions that are embedded in natural language provides a window onto how the brain organises affective experience at the individual level. Additionally, it gives a substrate for translational applications in mental health where clinically meaningful signals are precisely the inter-individual variability that population-average readouts miss.

\begin{figure}[!htbp]
  \centering
  \includegraphics[width=\textwidth]{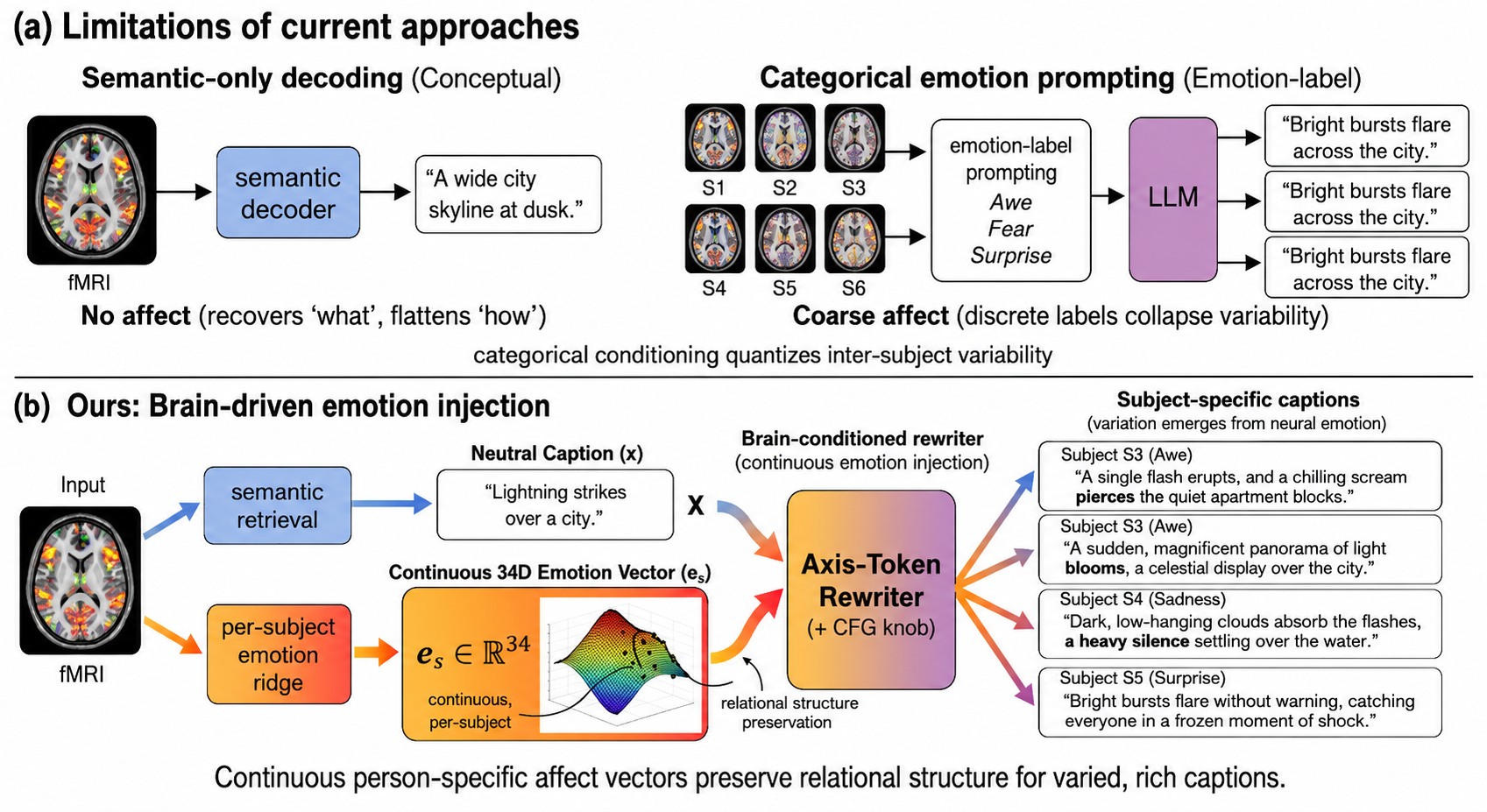}
  \caption{\textbf{(a)} Existing approaches decode the scene without
    affect~\citep{horikawa2024,tang2023,scotti2024} or condition LLMs
    on shared categorical labels~\citep{keskar2019,dathathri2020,sanchez2023}.
    \textbf{(b)~Ours.} Per-subject fMRI is decoded along two paths --
    semantic retrieval for the neutral caption $x$ and a per-subject
    ridge for the continuous 34-D Cowen \& Keltner vector $e_s$. The
    axis-token rewriter (Section~\ref{sec:rewriter}) maps $(x, e_s)$
    to a subject-specific affective caption.}
  \label{fig:pipeline}
\end{figure}


In recent years, deep learning has driven substantial progress in
generating diverse outputs from brain signals including reconstruction
of visual scenes~\citep{scotti2024,zangos2025},
speech~\citep{tang2023}, and linguistic
descriptions~\citep{horikawa2024,lu2025}. Recent
studies have further demonstrated open-vocabulary fMRI-to-text
generation by integrating LLMs directly into the decoding
pipeline~\citep{xi2023unicorn,chen2024bpgpt,du2023visual,li2022multiview,lin2022mindreader,duan2023dewave,cui2025brainx}.
However, as Figure~\ref{fig:pipeline}a illustrates, current approaches
still face limitations in handling the affective dimension of
perception. MindCaptioning trains a decoder to generate fluent
descriptions of perceived video content, but the supervision targets
are scene captions written without emotional
content~\citep{horikawa2024}. The semantic decoder
of~\citet{tang2023} reconstructs continuous perceived or imagined
speech from cortical activity, but also without an affect-conditioned
output channel. UniCoRN couples a brain encoder to a frozen LLM and
produces open-vocabulary text from fMRI, but the generation objective
is purely semantic~\citep{xi2023unicorn}. BrainX similarly aligns
brain features with a multimodal language model and shows
cross-subject text generation while remaining
affect-blind~\citep{cui2025brainx}. Controlled generation methods such
as CTRL and PPLM can steer LLM outputs toward emotion but only via a
small set of categorical control codes, which collapses inter-subject
variability into a discrete
set~\citep{keskar2019,dathathri2020,sanchez2023}. Therefore, it is essential to incorporate
insights from how the brain actually represents affect into the design
of brain-conditioned emotional generation models.

Neural representations of affect in the human brain are distributed,
high-dimensional, and individually variable. Decades of affective
neuroscience have shown that emotional experience is encoded in
distributed activation patterns spanning cortical and subcortical
networks~\citep{kragel2015,saarimaki2016}. These patterns vary
continuously rather than partitioning into discrete
categories~\citep{cowen2017} and individual subjects produce
distinct neural signatures even for identical naturalistic
stimuli~\citep{horikawa2020,lettieri2019}. A continuous high-dimensional vector decoded from brain activity can
in principle preserve these three properties. A top-5 categorical
emotion label cannot.

We present the first brain-driven emotional captioning pipeline. As
shown in Figure~\ref{fig:pipeline}b, per-subject fMRI is decoded along
two parallel streams. A neutral scene description is first obtained from brain-decoded language features
using retrieval-based brain captioning approaches~\citep{horikawa2024,lewis2020rag}. 
This caption is then paired with a continuous 34-dimensional emotion vector decoded by a 
subject-specific ridge model. The caption-emotion pair is passed to a trained rewriter based on an
encoder-decoder language model~\citep{lewis2020bart}, which uses learned emotion-axis conditioning
inspired by style-token and control-code methods~\citep{wang2018,keskar2019}. The rewriter produces
an emotion-styled caption while allowing inference-time control over the balance between semantic
content and affect via classifier-free guidance trained with an identity-preserving null branch.

Our contributions are: (i) the first brain-driven emotional captioning
pipeline conditioned on a continuous decoded emotion vector; (ii) a
three-axis validation framework (subject-specificity, structural
geometry, causal control) augmented by a TRIBE v2 substitution probe
that diagnoses representational gaps in current video-to-fMRI
encoders; (iii) head-to-head against strongly prompted GPT-4 across
two independent fMRI datasets, with ours dominating on all three axes.
\section{Related works}

\paragraph{Brain decoding from fMRI.}
Prior work decodes both semantic content from naturalistic fMRI using
retrieval-based and generative
pipelines~\citep{tang2023,horikawa2024,lu2025,scotti2024},
and emotion as categorical or dimensional variables from distributed
neural
activity~\citep{horikawa2020,kragel2015,kim2017,saarimaki2016,lettieri2019}.
The first line recovers textual or semantic embeddings but focuses on
scene content rather than emotional framing; the second treats emotion
as a target for analysis, not as a control signal for generation. Our
work preserves semantic decoding while using decoded emotion as the
conditioning signal that drives caption generation directly.

\paragraph{Controllable text generation.}
Machine learning methods for controllable generation steer sentiment,
style, or emotion through prompts, control codes, or classifier-free
guidance~\citep{mathews2016,gan2017,keskar2019,dathathri2020,krause2021,hosalimans2021,sanchez2023}.
These methods assume the control signal is externally specified by the
user or task~\citep{shen2017,prabhumoye2018}. In contrast, our setting
derives the control signal from decoded neural activity, making the
problem one of brain-conditioned control rather than user-conditioned
generation.

\paragraph{Prompting LLMs with emotion.}
LLMs are highly responsive to emotional content in prompts.
EmotionPrompt~\citep{li2023emotionprompt} shows that injecting emotional
stimuli into instructions yields measurable gains across 45 tasks, and
prompt-sentiment framing similarly modulates model
behaviour~\citep{gozzi2025framing}. These results establish text-based
emotion conditioning as a credible baseline for affective generation,
yet all such methods specify the conditioning signal externally as
labels or text descriptions and cannot read out a target subject's
continuous, person-specific affective state. Our strongest baseline prompts GPT-4 with brain-decoded top-5
C\&K labels (Section~\ref{sec:setup}). Our continuous
brain-conditioned pipeline is benchmarked against it.

\paragraph{Evaluation and the comparative gap.}
Prior controllable-generation work evaluates attribute fidelity and
content preservation~\citep{shen2017,prabhumoye2018}. Neuroscience
evaluates representational structure through similarity-based
analyses~\citep{kriegeskorte2008}. Either tradition alone is
insufficient for brain-conditioned generation. Attribute-fidelity
metrics show whether captions express target emotion but not whether
they reflect underlying neural geometry. Representational similarity
tests for that geometry but not whether the conditioning signal
causally drives generation. We combine the two and add subject-specificity
and modality-independent generalization under video-to-brain
substitution~\citep{dascoli2026}. No prior study has run a head-to-head
comparison of continuous brain-decoded conditioning against label-based
prompting on the same backbone and stimuli and we close this gap by
running both protocols on matched stimuli with the same evaluation
suite.

\section{Method}
Given the fMRI response of a subject viewing a held-out video clip, the
model produces a single caption that describes the scene and reflects
the subject's affective response to it (Figure~\ref{fig:pipeline}). The
input is a per-clip vector of voxel activations $\mathbf{v} \in
\mathbb{R}^{N_\mathrm{vox}}$ recorded during the clip, where
$N_\mathrm{vox}$ is the number of cortical voxels in standard MNI
space. The output is a
generated caption $\hat{y} = \{w_1, \ldots, w_n\}$ with each token
$w_i \in V$ over an open vocabulary. The mapping factors through two
intermediate representations decoded from $\mathbf{v}$: a neutral scene
description $x$ retrieved from a training caption corpus and a
34-dimensional Cowen \& Keltner emotion vector $e$ recovered by a
per-subject ridge (Section~\ref{sec:braindec}). A trained rewriter then
maps $(x, e)$ to $\hat{y}$ with inference-time control over how
strongly $e$ reshapes $x$ (Section~\ref{sec:rewriter},
Figure~\ref{fig:architecture}). The implementation details of each
component are described in the following subsections.

\subsection{Brain decoding}
\label{sec:braindec}

Stage~1 factors each subject's voxel response into two
complementary representations produced by per-subject linear decoders
fit on training clips and applied to held-out clips at test time.
Decoupling content from affect lets a single rewriter consume both
signals while preserving subject-specific magnitudes in the affect
channel (Section~\ref{sec:rewriter}, Figure~\ref{fig:pipeline}).

\textbf{Layer-wise Caption Retrieval.}
To recover the scene content $x$ from voxel signals, we map fMRI to a
multi-layer language-model feature stack and retrieve the
nearest-neighbor caption from a pre-indexed training corpus. Following
MindCaptioning~\cite{horikawa2024}, we train per-subject linear ridges
that map a subject's voxel response $\mathbf{v}$ to the mean-pooled
hidden state $h^{(\ell)} \in \mathbb{R}^{1024}$ at each of the 24
transformer layers of a pretrained DeBERTa-large language model. Voxel
features are first reduced via PCA because the high-dimensional voxel
space is much larger than the number of training clips and would
otherwise leave the ridge severely underdetermined. Using all 24
layers rather than just the final one captures DeBERTa's multi-scale
semantic structure and improves retrieval quality~\cite{horikawa2024,
tang2023}. At test time we predict a 24-layer feature stack from
held-out voxels and retrieve the nearest-neighbor caption $x$ in a
single pass:
\begin{equation}
x = \arg\max_{x' \in \mathcal{C}_{\rm train}}
\cos\!\bigl(\hat{\phi}(\mathbf{v}),\; \phi(x')\bigr),
\label{eq:retrieval}
\end{equation}
where $\hat{\phi}(\mathbf{v})$ is the brain-predicted concatenated
feature stack and $\phi(x')$ is the same stack pre-computed on a
candidate training caption (indexing details in
Appendix~\ref{app:retrieval}). This single-pass cosine lookup replaces
MindCaptioning's iterative per-token mask-fill, reducing inference
cost by orders of magnitude.

\textbf{Per-Subject Emotion Decoder.}
To produce the continuous emotion vector $e$ that conditions Stage~2,
we fit a per-subject linear decoder from fMRI voxels to the
34-dimensional Cowen \& Keltner emotion
taxonomy~\cite{cowen2017,horikawa2020}. The taxonomy was derived from
large-scale human ratings of evocative video clips and consists of 34
fine-grained subjective emotional states (e.g.\ Joy, Fear, Aesthetic
Appreciation, Romance, Empathic Pain, Calmness, Awe, Amusement,
Anxiety, Entrancement, Sadness, Disgust, Adoration, Nostalgia, Pride,
Triumph; full list in Appendix~\ref{app:hparams}), where each emotion is 
annotated as a continuous probability $\in[0,1]$ for every clip. We use this taxonomy
because it is the only continuous, fine-grained emotion space with
stimulus-paired annotations on the naturalistic film clips used in
our fMRI studies. Targets are the 34-dimensional ratings collected
from human annotators on the same clips, z-normalized per-dimension
and shared across subjects (the ratings are stimulus properties, not
subject responses; the per-subject ridge $W_s^\star$ carries the
individual differences). For each subject $s$ we reduce voxel
features to a low-dimensional PCA subspace
$\mathrm{PCA}_s(\mathbf{v})$ and fit a ridge regression to the
34-dimensional target $e_t^{\rm CK34}$ on the training-set clips
$t \in \mathcal{T}_{\rm train}$:
\begin{equation}
W_s^\star = \arg\min_{W}
\sum_{t \in \mathcal{T}_{\rm train}}
\bigl\| W\,\mathrm{PCA}_s(\mathbf{v}_t) - e_t^{\rm CK34} \bigr\|^2
+ \alpha \| W \|^2,
\qquad
e = W_s^\star\,\mathrm{PCA}_s(\mathbf{v}).
\label{eq:emodec}
\end{equation}
At test time the per-subject ridge predicts a 34-dimensional emotion
vector $e$ for each held-out clip, which becomes the conditioning
input to Stage~2. PCA dimensionality and $\ell_2$ penalty are reported
in Appendix~\ref{app:hparams}.
\subsection{Axis-token rewriter}
\label{sec:rewriter}

\begin{figure}[!htbp]
\centering
\includegraphics[width=\linewidth]{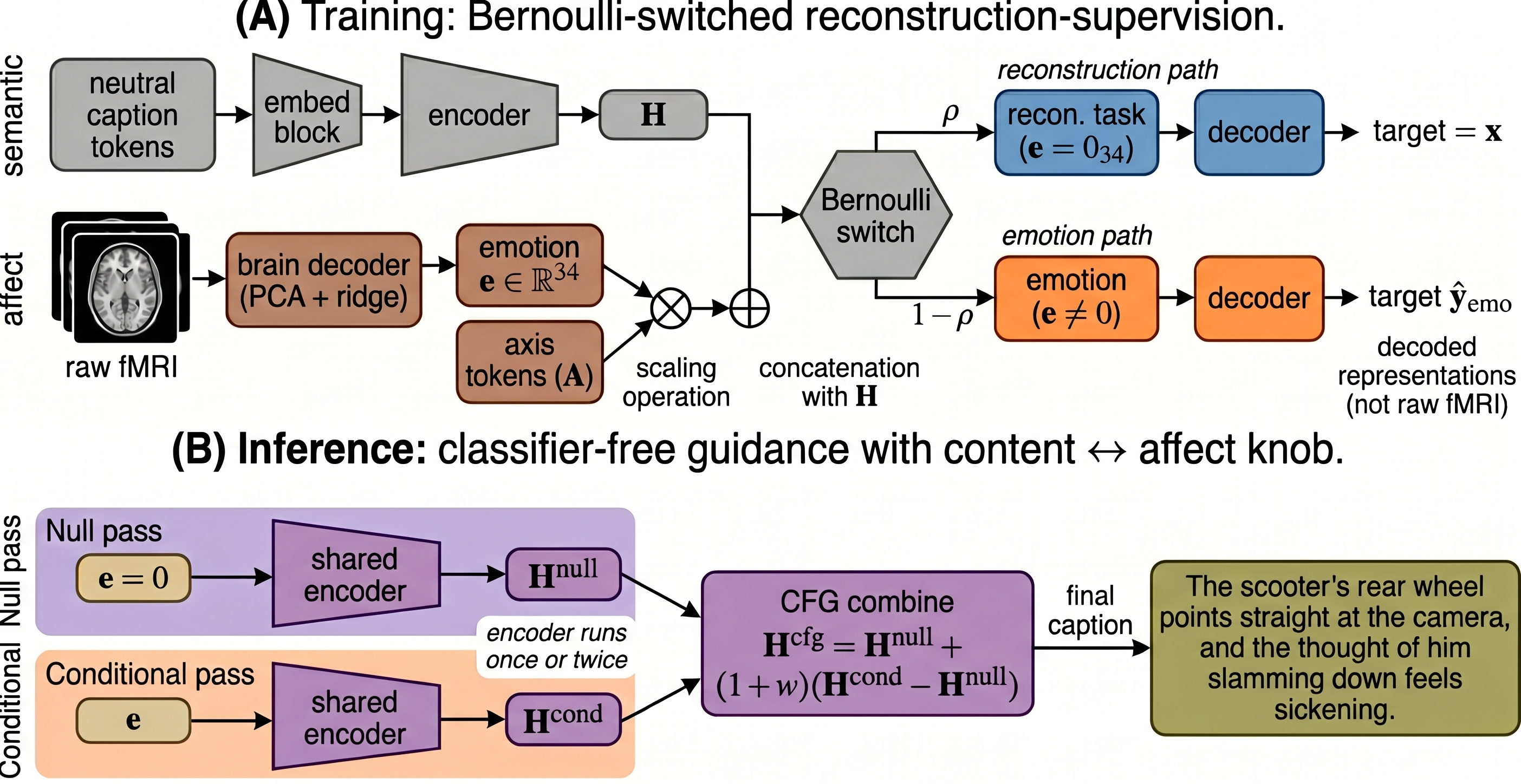}
\caption{\textbf{Axis-token rewriter.}
\textbf{(A)~Training} mixes a reconstruction target ($\mathcal{L}_{\rm recon}$,
probability $\rho$) and an emotion-specific target
($\mathcal{L}_{\rm emo}$) via a Bernoulli switch on $e\in\mathbb{R}^{34}$,
which scales axis matrix $A\in\mathbb{R}^{34\times 768}$. \textbf{(B)~Inference}
applies classifier-free guidance, $H^{\rm cfg}=H^{\rm null}+(1{+}w)(H^{\rm cond}-H^{\rm null})$,
trading content fidelity (smaller $w$, $w{=}0$ recovers $H^{\rm cond}$) for amplified affect (larger $w$).}
\label{fig:architecture}
\end{figure}

The rewriter (Figure~\ref{fig:architecture}) maps the neutral caption
$x$ and brain-decoded 34-D emotion vector $e$ from
Section~\ref{sec:braindec} to an emotion-styled caption $\hat{y}$.

\textbf{Architecture.}
The rewriter is a pretrained BART-base encoder-decoder
($d{=}768$)~\cite{lewis2020bart} augmented with 34 learned axis
tokens, jointly fine-tuned on the available training corpus. The BART
encoder produces a hidden state $H \in \mathbb{R}^{T \times d}$ from
the input caption. A learned matrix $A \in \mathbb{R}^{34 \times d}$
holds one vector per Cowen \& Keltner category. At
each forward pass, the brain-decoded emotion vector $e$ scales each
axis vector element-wise to yield the conditional encoder output:
\begin{equation}
H^{\mathrm{cond}} = \mathrm{concat}\!\left(H,\; e \odot A\right) \in
\mathbb{R}^{(T+34) \times d},
\label{eq:cond}
\end{equation}
where $(e \odot A)_k = e_k A_k$ is the $k$-th axis vector scaled by
the $k$-th emotion intensity. The null encoder output uses a zeroed
emotion: $H^{\mathrm{null}} = \mathrm{concat}(H, \mathbf{0})$. The BART
decoder cross-attends to either $H^{\mathrm{cond}}$ or
$H^{\mathrm{null}}$ during training and the two are linearly combined
at inference as shown below. Concatenating scaled axis tokens to the encoder output, 
instead of injecting them into decoder cross-attention or prepending them as soft 
prompts, preserves the pretrained encoder representation of $x$ while allowing the 
decoder to interpret the emotion axes as additional virtual context tokens.

\textbf{Training Objective: Bernoulli-switched Reconstruction-supervision.}
To make inference-time guidance interpolate between content
preservation and emotion override, rather than between unconditional
and conditional generation as in standard CFG, we train the rewriter
with a Bernoulli-switched objective that mixes a reconstruction-identity task
and an emotion-target task within a single model.
Given a triple $(x, e, y_{\mathrm{emo}})$ of a neutral caption $x$,
brain-decoded emotion $e$, and paired emotion-specific caption
$y_{\mathrm{emo}}$ for the same clip, the loss is:
\begin{equation}
\mathcal{L}(x, e, y_{\mathrm{emo}}; \rho) =
\rho \cdot \mathrm{CE}\!\left(f_\theta(x, \mathbf{0}_{34}),\; x\right) +
(1 - \rho) \cdot \mathrm{CE}\!\left(f_\theta(x, e),\; y_{\mathrm{emo}}\right),
\label{eq:loss}
\end{equation}
where $f_\theta(x, e)$ is the rewriter conditioned on emotion vector
$e$, $\mathbf{0}_{34} \in \mathbb{R}^{34}$ is the null emotion vector
(typographically bolded to distinguish from the scalar zero in $\rho \in [0,1]$),
and $\rho \in [0, 1]$ is the reconstruction probability. With probability $\rho$ the 
null condition is used and the target is the input caption itself. The full emotion 
vector conditions the model with probability $1-\rho$, in which case the target is 
the corresponding emotion-specific caption. This extends standard
condition-dropout training for classifier-free guidance with an
explicit identity target for the null branch. Standard CFG learns an
unconditional generation distribution in the null pathway and
interpolates between unconditional and conditional generation at
inference. Our formulation instead learns input reproduction in the
null pathway, so guidance at inference interpolates directly between
semantic fidelity and emotion-conditioned override. Because decoded
neural affect is noisy and variable across subjects, this
identity-preserving null branch is essential since it teaches the model
when to preserve scene content and when to allow affective control to
reshape the caption. Training data and optimization details are in
Section~\ref{sec:setup} and Appendix~\ref{app:hparams}.

\textbf{Inference: Classifier-free Guidance.}
To expose a single inference-time knob that controls how strongly
decoded affect reshapes the output, we apply classifier-free guidance
over the conditional and null encoder outputs.
At inference, we combine the conditional and null encoder outputs of
Equation~\ref{eq:cond} into a single guided encoder output and run the
BART decoder once:
\begin{equation}
H^{\mathrm{cfg}} = H^{\mathrm{null}} + (1 + w)\bigl(H^{\mathrm{cond}} -
H^{\mathrm{null}}\bigr),
\label{eq:cfg}
\end{equation}
where $w \geq 0$ is the guidance weight. Rearranging gives
$H^{\mathrm{cfg}} = (1+w) H^{\mathrm{cond}} - w H^{\mathrm{null}}$, i.e.\
the guided encoding is a linear extrapolation along the affect direction
that subtracts $w$ copies of the neutral encoding from $(1{+}w)$ copies
of the emotion-conditional one. At $w=0$ the output recovers the
standard conditional encoding $H^{\mathrm{cond}}$ (no extrapolation), and at
$w>0$ the model amplifies affective expression by moving further along the
$H^{\mathrm{cond}} - H^{\mathrm{null}}$ direction, away from the
reconstruction-supervised null. Larger $w$ therefore trades content
fidelity for stronger affect. While CFG is standard in image
diffusion, applying it to a BART encoder hidden state is non-standard
here because it requires the conditional and null encodings to live in the
same continuous space. Axis-token concatenation guarantees
since both pathways share the BART encoder weights and differ only in
the 34 trailing tokens.

\section{Experiments}
\label{sec:experiments}

\subsection{Experimental setup}
\label{sec:setup}

\paragraph{Datasets.}
\textbf{MindCaptioning (MC)}~\cite{horikawa2024}: 6 subjects, 2{,}108 short
video clips, 3T fMRI, up to 20 crowd-written neutral captions per clip;
canonical 2{,}036 train / 72 test split. \textbf{Horikawa 2020
(H2020)}~\cite{horikawa2020}: 5 subjects, 3T fMRI to 2{,}181 short
emotional video clips with 34-D Cowen \& Keltner~\cite{cowen2017}
category ratings; used only for cross-dataset evaluation.
\textbf{Emo-FilM}~\cite{emofilm2024}: 9 long-form films with 50-D
continuous per-second emotion annotations, used only as the
synthetic-brain validation substrate. The Stage-2 rewriter is additionally
trained on a set of LLM-generated emotion-targeted caption corpus over the MC \emph{training} clips only
(the 72 MC test clips were excluded from the GPT-4o rewriting prompts;
no test-clip text appears in Stage-2 training). Full prompt, counts,
and provenance caveats are in Appendix~\ref{app:capdata}.

\paragraph{Baselines.}
(i) \textbf{Retrieval-only}: Stage-1 caption, no rewriter.
(ii) \textbf{GPT-4 (OWN)}: GPT-4o prompted with the neutral caption and
the \emph{ground-truth} top-5 C\&K labels for the clip.
(iii) \textbf{GPT-4 (SWAP)}: identical to (ii) but with target labels
shifted by 36 clips, matching our SWAP protocol. GPT-4 receives stronger
symbolic supervision than our model but has no access to continuous,
subject-specific brain-decoded affect; the asymmetry distinguishes emotion
\emph{communication} from emotion \emph{induction from neural activity}.

\paragraph{Evaluation metrics.}
We evaluate all captions using our retrained \textbf{CK34 scorer}, a
DeBERTa-v3 regressor fine-tuned on Cowen \& Keltner's 2{,}185 video
annotations ($r=0.79$ held out). The scorer is applied identically to
our method and to all baselines.

Our primary fidelity metric is \emph{all-34 $r$}, defined as the mean
Pearson correlation between the brain-decoded 34-D emotion vector and
the CK34 score of the generated caption. We also report
\emph{semantic identification accuracy}, measured by cosine pairwise
rank with chance performance at $0.5$, and \emph{inter-subject
diversity}, computed as the pairwise distance among the six MC
subjects' captions for each clip using CK34 cosine distance, lexical
unigram distance, and character edit distance.

To test whether the rewriter is genuinely controlled by the conditioning
emotion vector, we introduce \emph{SWAP own-leakage}. In the SWAP setting,
the model receives a target emotion vector from a different clip. We then
measure the fraction of generated captions whose CK34 top-1 emotion still
matches the original, un-swapped clip's top emotion. A model that fully
follows the swapped target should leak at chance
($1/34{\approx}2.9\%$). One that falls back on the input scene's affect
leaks substantially more.

We further report \emph{RSA Spearman $\rho$}, the Spearman correlation
between the brain-decoded RDM and the caption-CK34 RDM, computed over
test clips. Per-subject results are reported as mean $\pm$ s.d. We compare
our method against GPT-4 using paired Wilcoxon signed-rank tests by clip;
bootstrap confidence intervals for diversity metrics use $B=10{,}000$
resamples.

The two design knobs from Section~\ref{sec:rewriter}, the
reconstruction-supervision rate $\rho$ and the CFG weight $w$, are fixed
throughout to $\rho{=}0.4$ and $w{=}2$ (joint sweep in
Appendix~\ref{app:param}). Appendix-only ablations sweep around this
operating point.

Stage-1 ridge models are fit on CPU. The Stage-2 rewriter trains in
approximately 4 GPU-hours on a single A100 GPU with 40~GB memory. The
full project, including sweeps and ablations, required approximately
60--80 GPU-hours. Implementation details, including optimizer,
hyperparameters, and hardware, are provided in Appendix~\ref{app:hparams}.
Standard NLG metrics, including BLEU, ROUGE, and BERTScore, are reported
in Appendix~\ref{app:nlg}.

\begin{table}[!htbp]
\centering
\caption{Three-axis validation on MC ($n{=}6$ subjects, 72 test
clips). Mean $\pm$ s.d.\ across subjects. Bold marks the best between
Ours and GPT-4 (OWN). \emph{Per-clip CK34 cosine}: cosine between
brain-decoded 34-D vector and rewriter-output CK34 score, averaged
over $n{=}6{\times}72$. SWAP chance ${=}1/34{\approx}2.9\%$. \emph{Sem-id}
favours GPT-4 because GPT-4 receives the unmodified neutral caption as
input while ours maps from a noisy brain decode; this metric is reported
for completeness, not as a headline. TRIBE synthetic-brain probe in
Section~\ref{sec:tribe}; per-subject and per-axis decompositions in
Appendices~\ref{app:loso},~\ref{app:rsa}.}
\label{tab:main}
\small
\setlength{\tabcolsep}{6pt}
\begin{tabular}{p{2.4cm}lcc}
\toprule
Axis & Metric & Ours & GPT-4 (OWN) \\
\midrule
Subject-specificity     & CK34-cosine diversity        & $\mathbf{0.615}$           & $0.179$ \\
                        & Lexical-unigram diversity    & $\mathbf{0.809}$           & $0.070$ \\
                        & Character-edit diversity     & $\mathbf{0.645}$           & $0.058$ \\
\midrule
Structural (RSA)        & Spearman $\rho$ (group)      & $\mathbf{+0.635}$          & $+0.435$ \\
\midrule
Causal (SWAP)           & own-leakage $\downarrow$     & $\mathbf{2.8\%}$           & $8.3\%$ \\
                        & target $r$ effect            & $\mathbf{+0.427}$          & $+0.198$ \\
\midrule
Headline                & Per-clip CK34 cosine         & $\mathbf{+0.345 \pm 0.010}$ & $+0.240 \pm 0.204$ \\
                        & Sem-id (rewriter output)     & $0.791$                    & $\mathbf{0.998}$ \\
\bottomrule
\end{tabular}
\end{table}

We organise the analysis along three axes (Table~\ref{tab:main}). As a cross-cutting
robustness test, we also replace measured fMRI with synthetic-brain output from
\emph{TRIBE v2}~\cite{dascoli2026}. TRIBE v2 is a frozen, large-scale video-to-fMRI encoder 
trained on a naturalistic film-watching corpus. It predicts subjects' BOLD responses from 
video features alone with no fMRI input at test time. We pass the TRIBE-predicted voxel 
activations through our existing per-subject Stage-1 ridges without retraining. We then 
generate captions for 9 Emo-FilM films across all 6 MC ridges. This analysis isolates 
which findings depend on direct neural measurement and which can be reproduced by a synthetic
substitute. Full details are provided in Appendix~\ref{app:tribe}. Per-axis decompositions,
per-subject statistics, and statistical controls are provided in Appendices~\ref{app:rsa},
~\ref{app:swap2x2},~\ref{app:noise}.

\textbf{Axis 1 (Subject-specificity).}
The cleanest demonstration that decoded affect carries information
labels cannot is whether the same scene viewed by different subjects
produces \emph{different captions in proportion to their neural
differences}. We measure pairwise distance between the 6 subjects'
captions per MC test clip on three metrics (CK34 cosine, lexical
unigram, character Levenshtein) and compare against GPT-4 prompted
with each subject's brain-decoded top-5 labels. Per-subject brain
conditioning is $3.4\times$ more diverse on CK34 cosine, $11.6\times$
on lexical, and $11.1\times$ on character edit (ratios of group means in
Table~\ref{tab:main}, Axis~1; all $p < 10^{-11}$, paired Wilcoxon
signed-rank over $n{=}72$ clip-level paired differences). The gap is
largest on lexical and character-edit metrics that capture surface
form ($\sim 11\times$) and smallest on CK34 cosine that captures
shared affect ($\sim 3\times$), the diversity grows where labels cannot
reach. Critically, the diversity is \emph{organised by neural
similarity} rather than random. Pooling all 1080 subject-pair points
across the 72 clips ($n{=}1080$), brain-pair similarity predicts
caption-pair similarity at $r_{\rm CK34}=+0.33$ (analytic Pearson
$p<10^{-29}$) with within-clip subject-swap permutation
($B{=}10{,}000$) yielding $p<10^{-4}$ and confirming the effect is not
a clip-level confound. A distribution-matched Gaussian noise
control (preserving the empirical $34\times34$ covariance) removes
most of this structure, which is shown by emotion alignment and representational geometry
collapsing by a factor ${\sim}4\times$. This indicates that diversity alone is not
diagnostic, and brain conditioning constrains variability into a structured
form aligned with neural similarity (Appendix~\ref{app:noise}).

\textbf{Axis 2 (Structural / RSA).}
This axis tests whether the brain-decoded conditioning carries
relational structure between clips that label-based prompting cannot
reproduce. We compute two cosine RDMs over the 72 test clips where one is from the brain-decoded 
34-D emotion vectors and the other is from the CK34 scores of the generated captions.
We then compare their upper triangles using Spearman $\rho$. At the group level
(Table~\ref{tab:main}, Axis~2), our method yields $\rho=+0.635$ while GPT-4 (GT
top-5) returns $+0.435$. At the per-subject level the architectural advantage
attenuates ($+0.343\pm0.037$ vs.\ GPT-4 brain top-5 $+0.334\pm0.030$),
indicating subject-specific structural alignment is carried by the
brain decode itself (full decomposition in Appendix~\ref{app:rsa}).

\textbf{Axis 3 (Causal / SWAP).}
For each test clip we generate \emph{OWN} (target $=$ clip's
brain-decoded vector) and \emph{SWAP} (target $=$ vector shifted by 36
in the test list) captions. We then measure (a) per-dim Pearson $r$
between target and CK34 score (averaged across 34 dims) and (b)
own-leakage as defined in Section~\ref{sec:setup}. Our model attains
$r_{\rm target}-r_{\rm own}=+0.427$ and own-leakage $\mathbf{2.8\%}$, which are
statistically indistinguishable from the $1/34{\approx}2.9\%$ chance
rate. This indicates the rewriter follows the supplied target without residual
scene-affect bleed-through. GPT-4 (SWAP), prompted with shifted GT top-5
labels, attains +0.198 and 8.3\% (${\sim}3\times$ above chance). Continuous
brain conditioning therefore exceeds clean categorical conditioning even
with ground-truth target labels.

\subsection{Diagnosing structural-geometry loss in synthetic-brain encoders}
\label{sec:tribe}

\begin{figure}[!htbp]
\centering
\includegraphics[width=\linewidth]{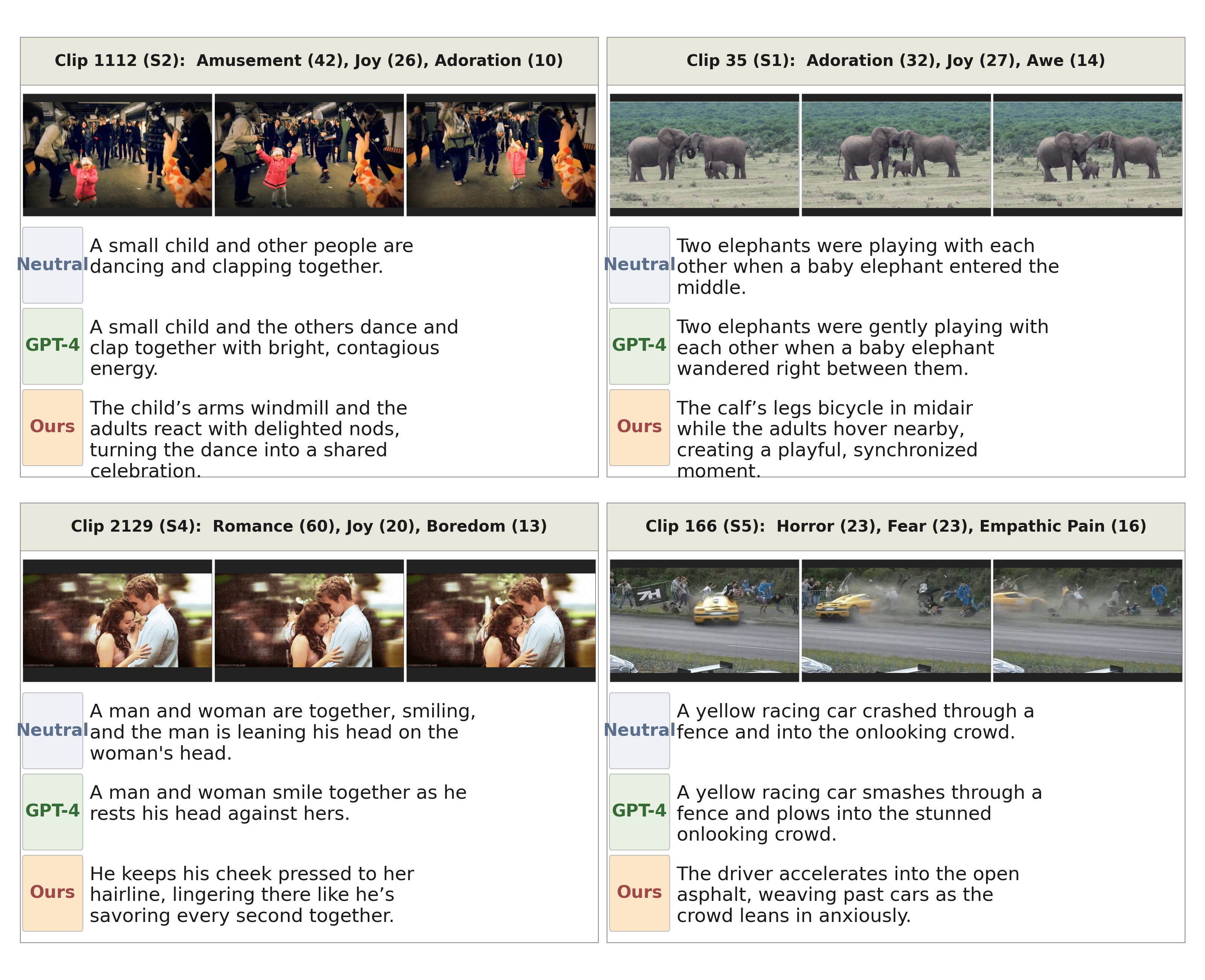}
\caption{\textbf{Qualitative captions across four affect regimes
(Entrancement, Nostalgia, Romance, Fear/Horror).} Each cell, one MC
test clip with its dominant CK34 emotions, three frames, and three
captions. \emph{Neutral} is Stage-1 retrieval (no affect).
\emph{GPT-4} is GPT-4o prompted with the clip's GT top-5 C\&K labels.
\emph{Ours} is the brain-conditioned rewriter for one subject. The
Fear cell illustrates the gap. GPT-4 describes what the camera
records, whereas ours captures what the viewer feels. Additional
examples in Appendix~\ref{app:samples}.}
\label{fig:qualitative}
\end{figure}

To probe what current voxel-prediction models retain about affect, we
substitute measured fMRI with synthetic activity from \emph{TRIBE
v2}~\cite{dascoli2026}, the current SOTA video-to-fMRI encoder, to
predict each subject's fMRI from video features. We pass
TRIBE's predicted activity through our six per-subject ridges to
obtain synthetic 34-D affect vectors, then through our rewriter model
(Section~\ref{sec:rewriter}). Two metrics dissociate sharply.

\emph{SWAP own-leakage} (Axis~3) measures whether a synthetic-brain
substitute still drives the rewriter to follow a shifted target rather
than reverting to the input clip's affect. Leakage rises only to
$5.2\%\pm1.8\%$ from the real-fMRI value of $2.8\%$, still well below
GPT-4 with ground-truth labels ($8.3\%$). Point-wise affect content is
preserved, showing that TRIBE-predicted activity carries enough clip-specific
signal for our ridges to recover a usable target vector.

\emph{RSA} (Axis~2) measures whether the between-clip distance
structure decoded from brain matches the structure of the generated
captions. The Spearman correlation drops from $\rho{=}{+}0.635$ with real fMRI to
$\rho{=}{+}0.166$ with TRIBE-substituted activity, retaining only
${\sim}26\%$ of the original magnitude. This indicates that much of the
relational structure is lost. Synthetic activity preserves information
about the affect evoked by each clip individually, but does not preserve
how clips are differentiated from one another in affective space.
The dissociation isolates a representational gap not visible in the
SOTA encoder's own benchmark scores. RSA between real and
synthetic fMRI is a stricter generalization criterion than per-clip
prediction accuracy for next-generation brain encoders.

\subsection{Cross-dataset transfer and comparison with prior decoders}
\label{sec:compare}

On \textbf{H2020}~\cite{horikawa2020} ($n{=}5$; different cohort,
stimuli, preprocessing), the MC-trained Stage-2 rewriter applied
unchanged on per-subject H2020 Stage-1 ridges attains all-34
$r = 0.253 \pm 0.026$, comparable to within-MC $0.265 \pm 0.027$
(Table~\ref{tab:main}; Appendix~\ref{app:h2020}).

Table~\ref{tab:prior} compares our method with prior brain-to-text
decoders on the same 72 MC test clips. Because prior methods target descriptive recall 
rather than emotion conditioning, their all-34 $r$ falls to chance. Our model preserves
semantic identification while adding the affective channel prior decoders
leave unmodeled.

\begin{table}[t]
\centering
\caption{Comparison with prior brain-to-text decoders on MC test clips.
\emph{Sem-id}: pairwise-rank semantic identification accuracy.
\emph{all-34 $r$}: brain-emotion fidelity. Prior decoders do not
condition on emotion; all-34 $r$ near zero is expected and is
the gap our model fills.}
\label{tab:prior}
\small
\begin{tabular}{lccc}
\toprule
Method & sem-id & all-34 $r$ & emotion-conditioned \\
\midrule
MindCaptioning iterative~\cite{horikawa2024} & $0.642$ & $\sim 0$ & no \\
Retrieval-only (Stage 1) & $\mathbf{0.951}$ & $\sim 0$ & no \\
GPT-4 (OWN, GT top-5) & --- & --- & yes (categorical) \\
\textbf{Ours} (Stage 1 + axis-token rewriter) & $\mathbf{0.951}$ & $\mathbf{0.265 \pm 0.022}$ & yes (continuous) \\
\bottomrule
\end{tabular}
\end{table}

Surface NLG metrics favour GPT-4 (BLEU-4 $39.0$ vs.\ $1.9$) because
it has access to ground-truth labels (full table in
Appendix~\ref{app:nlg}). The affect-fidelity ranking reverses on
Table~\ref{tab:main} once measured against what GPT-4 lacks access to.

\section{Discussion}
\label{sec:discussion}
The three-axis evaluation supports our comparative thesis. The
strongest evidence is the SWAP setting, where the brain-conditioned
rewriter follows shifted affective targets at chance-level own-leakage
while label-prompted GPT-4 leaks the original clip's affect through.
Across all three axes, our model outperforms label-prompted GPT-4
wherever subject-specific or relational affective structure matters,
and the inter-subject diversity is organised by neural similarity rather
than arising as uncontrolled variation. The dissociation is not about
general caption quality (GPT-4 leads on surface-overlap metrics and
matches per-subject RSA when given the same brain decode) but about
information content. Continuous neural-affect vectors carry relational
and subject-specific structure that categorical emotion labels cannot
transmit.

The TRIBE substitution analysis (Section~\ref{sec:tribe}) further
separates causal usability from representational structure. Point-wise
affect content is preserved through synthetic activity, but much of the
between-clip structure is lost, suggesting affective content depends on
interoceptive, contextual, or subject-specific signals not yet captured
by current video-to-fMRI encoders. The two-stage decomposition keeps
semantic and affective decoding on separate channels, positioning
individual brain-decoded affect as a substrate for studying how affect
is computed separately from semantics, and as a person-specific signal
that population-average label decoders tend to erase.

\emph{Limitations.} The pipeline requires per-subject calibration with
matched-stimulus fMRI, produces a single 34-D vector per clip
(collapsing within-clip temporal dynamics), and is bounded by the
Cowen\,\&\,Keltner 34-D taxonomy.

\section{Conclusion}
\label{sec:conclusion}

Decoded neural affect transmits subject-specific relational structure
that categorical-label conditioning cannot recover and synthetic-brain
substitution shows causal usability transfers across measurement
apparatus while structural usability is partly apparatus-bound. The
signal is worth pursuing wherever subject-specific perceptual
experience matters.

\bibliographystyle{unsrtnat}
\bibliography{references}

\newpage
\appendix
\section{Appendix}

\subsection{Retrieval implementation details}
\label{app:retrieval}

We pre-compute the 24-layer DeBERTa feature stack for every caption in
the training corpus once and store the concatenated $24 \times
1024$-dimensional index vector
$\phi(x') = [h^{(1)}(x');\ldots;h^{(24)}(x')]$. At test time we pass
held-out fMRI voxels through the 24 ridges in parallel to obtain a
brain-predicted feature stack
$\hat{\phi}(\mathbf{v}) = [W^{(1)}\mathbf{v};\ldots;
W^{(24)}\mathbf{v}]$, and apply Equation~\ref{eq:retrieval}. The
single-pass cosine lookup is what makes the thousands-of-runs
three-axis evaluation tractable; without it, the GPT-4-comparison and
TRIBE-substitution runs would not be feasible at the scale this paper
reports. In evaluations that isolate Stage~2 (the GPT-4 baseline and
the synthetic-brain generalization runs), we substitute a
ground-truth training caption from the clip's paired annotations for
$x$, removing Stage~1 variance as a confound.

\subsection{Hyperparameters}
\label{app:hparams}

\begin{table}[h]
\centering
\small
\begin{tabular}{lll}
\toprule
Component & Hyperparameter & Value \\
\midrule
Brain-decoded $x$ (retrieval ridges) & PCA components & 500 \\
                                     & Ridge $\ell_2$ penalty $\alpha$ & $10^{4}$ (per-layer) \\
Brain-decoded $e$ (emotion ridge)    & PCA components & 500 \\
                                     & Ridge $\ell_2$ penalty $\alpha$ & 100 \\
Axis-token rewriter                   & Backbone & BART-base ($d{=}768$) \\
                                     & Axis matrix $A$ & $\mathbb{R}^{34\times 768}$, random init \\
                                     & Reconstruction probability $\rho$ & 0.4 \\
                                     & Optimizer & AdamW \\
                                     & Learning rate & $5 \times 10^{-5}$ \\
                                     & Schedule & constant \\
                                     & Epochs & $\sim$5 \\
                                     & Batch size & 32 \\
Inference                            & CFG weight $w$ & $2$ (default) \\
\bottomrule
\end{tabular}
\caption{Hyperparameters for all stages of the pipeline.}
\end{table}

\subsection{Parameter sensitivity ($\rho$, $w$)}
\label{app:param}

\begin{figure}[h]
\centering
\includegraphics[width=\linewidth]{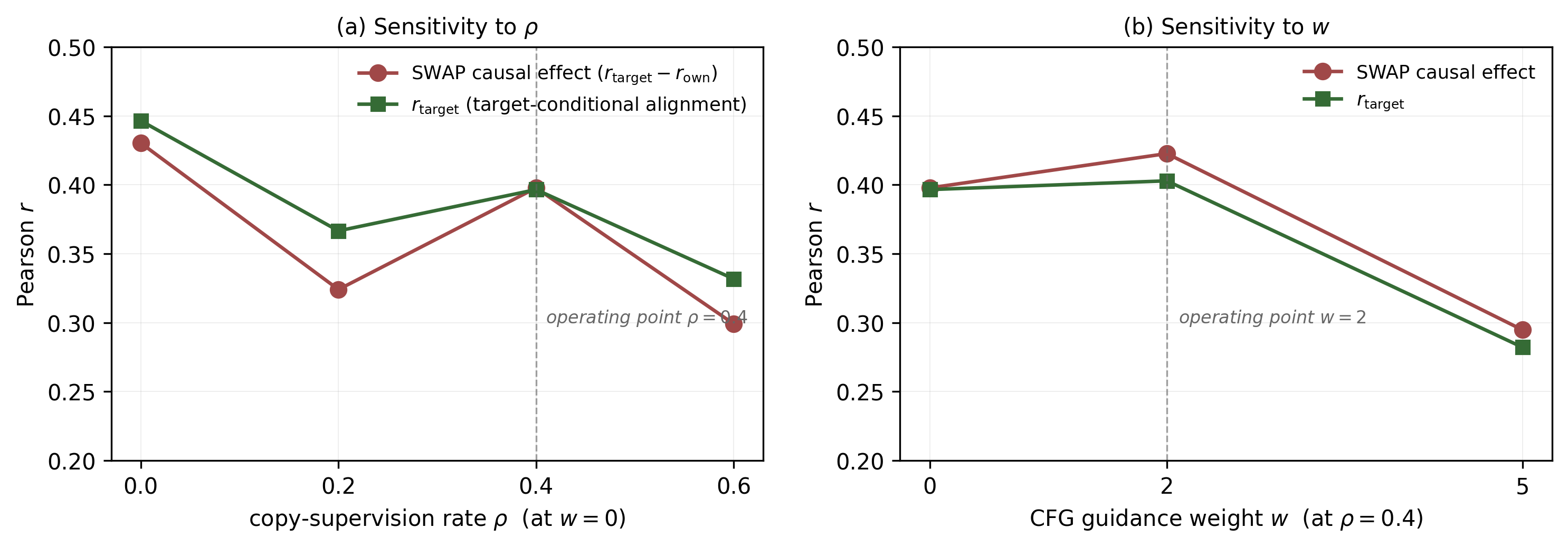}
\caption{\textbf{Parameter sensitivity.} \textbf{(a)} SWAP causal
effect and target-conditional alignment $r_{\rm target}$ as functions
of reconstruction-supervision rate $\rho$ (with $w{=}0$ fixed). \textbf{(b)}
Same metrics as functions of CFG guidance weight $w$ (with
$\rho{=}0.4$ fixed). Both metrics remain positive across the full
range tested. $\rho{=}0.4$, $w{=}2$ marks the operating point used
throughout the paper.}
\label{fig:param}
\end{figure}

The two control knobs of Section~\ref{sec:rewriter}---reconstruction-supervision
rate $\rho$ and classifier-free guidance weight $w$---are the
load-bearing hyperparameters of the design. We sweep each
independently (Figure~\ref{fig:param}), holding the other at the
operating point ($\rho{=}0.4$, $w{=}2$). The SWAP causal effect
($r_{\rm target} - r_{\rm own}$) and target-conditional fidelity
($r_{\rm target}$) remain positive across the entire range tested
($+0.30$ to $+0.45$). $\rho{=}0.4$ balances the reconstruction-identity and
emotion-target tasks. At $\rho{=}0$ the model has no identity
grounding and stylistic content dominates, while $\rho{=}0.6$
over-weights reconstruction and erodes the emotion target. $w{=}2$ similarly
sits at the inflection point. $w{=}0$ recovers the standard conditional
encoding (no CFG amplification, lower causal effect) and $w{=}5$
extrapolates too aggressively, degrading both metrics.

\subsection{Cross-dataset transfer detail (Horikawa 2020)}
\label{app:h2020}

Per-subject all-34 $r$ for the $n=5$ H2020 subjects, evaluating the
MC-trained pipeline with no fine-tuning. Mean all-34 $r = 0.253 \pm 0.026$;
training-defined top-10 $r = 0.200 \pm 0.055$ (random-10 baseline 0.252).

\begin{table}[h]
\centering
\small
\begin{tabular}{lccc}
\toprule
Subject & sem (rewrite) & all-34 $r$ & top-10 $r$ (TRAIN) \\
\midrule
S1   & 0.873 & 0.216 & 0.114 \\
S2   & 0.891 & 0.246 & 0.198 \\
S3   & 0.866 & 0.264 & 0.214 \\
S4   & 0.905 & 0.253 & 0.205 \\
S5   & 0.852 & 0.288 & 0.267 \\
\midrule
mean & 0.877 & 0.253 & 0.200 \\
s.d. & --- & 0.026 & 0.055 \\
\bottomrule
\end{tabular}
\end{table}

\subsection{Qualitative caption examples}
\label{app:samples}

Six MC test clips spanning four wins, one mediocre, and one failure case.
\emph{Top-3 emotions} are scene-level Cowen \& Keltner ratings (percentage
votes from the original C\&K human annotators). Ours and GPT-4 condition
on the same dominant emotions; the contrast is in how each model
expresses them. The bottom failure case is a faithful rendering of the
brain decoder's error, not a rewriter failure.

\begin{table}[h]
\small
\begin{tabularx}{\linewidth}{l X}
\toprule
\multicolumn{2}{l}{\textbf{Clip 715} --- Amusement (62), Awe (13), Interest (13)} \\
Ours      & The baby's legs bicycle in midair while the adult holds firm, making the struggle feel like a shared victory lap. \\
GPT-4     & A baby monkey tries to scamper away, but the adult monkey casually catches him by the leg. \\
Reference & Baby monkey tries to escape, but the adult monkey grabs him by the leg. \\
\midrule
\multicolumn{2}{l}{\textbf{Clip 1112} --- Amusement (42), Joy (26), Excitement (10)} \\
Ours      & The child's arms windmill and the men laugh it off, turning the dance into a shared celebration. \\
GPT-4     & A small child and the others dance and clap together with bright, contagious energy. \\
Reference & A small child and other people are dancing and clapping together. \\
\midrule
\multicolumn{2}{l}{\textbf{Clip 1295} --- Romance (44), Adoration (25), Joy (19)} \\
Ours      & He leans in again, kissing her gently, and she tilts her head back as if savoring the closeness. \\
GPT-4     & A woman lies in the garden while a man leans over her and kisses her softly on the nose. \\
Reference & A woman is lying in the garden, and another man is on top of her, giving her a kiss on her nose. \\
\midrule
\multicolumn{2}{l}{\textbf{Clip 499} --- Entrancement (38), Aesthetic Appreciation (31), Calmness (19)} \\
Ours      & The alternating black and white bands form a minimalist sculpture, each wave matching the last as it rotates. \\
GPT-4     & White and black waves ripple in sequence across the frame, slow and mesmerizing. \\
Reference & Two-dimensional waves of white and black color in a sequential wave format. \\
\midrule
\multicolumn{2}{l}{\textbf{Clip 1936} --- Craving (46), Aesthetic Appreciation (17), Satisfaction (8) \quad \emph{[mediocre]}} \\
Ours      & The camera lingers on the cob's glossy sheen, inviting a closer look at textures and timing. \\
GPT-4     & A pat of butter melts over a corn cob, glistening as it seeps into every crack. \\
Reference & A slice of butter is melting onto a corn cob and seeping into the cracks. \\
\midrule
\multicolumn{2}{l}{\textbf{Clip 1596} --- Empathic Pain (39), Disgust (39), Anxiety (5) \quad \emph{[failure]}} \\
Ours      & The camera lingers on the person's lips and tongue, inviting curiosity about what they'll do next. \\
GPT-4     & A pin is pushed into a person's tongue for a piercing. \\
Reference & A pin is being placed in a person's tongue for piercing. \\
\bottomrule
\end{tabularx}
\end{table}

\subsection{Standard NLG metrics}
\label{app:nlg}

GPT-4 wins decisively on surface-overlap metrics because it receives the
unmodified neutral caption and the ground-truth top-5 emotion labels as
input; our model receives only brain-decoded vectors. These metrics
were never the headline; they are reported here for completeness.

\begin{table}[h]
\centering
\small
\begin{tabular}{lcccccc}
\toprule
Variant & BLEU-1 & BLEU-2 & BLEU-3 & BLEU-4 & ROUGE-F & BERT-F \\
\midrule
Retrieval only (Stage 1) & 52.6 & 26.5 & 13.3 & 6.2  & 35.5 & 90.3 \\
Ours (brain S1)          & 34.7 & 12.1 & 4.3  & 1.3  & 23.3 & 87.6 \\
Ours (GT S1)             & 36.7 & 13.8 & 5.3  & 1.9  & 27.7 & 88.2 \\
Ours (SWAP)              & 32.2 & 10.6 & 3.6  & 1.2  & 23.6 & 86.7 \\
GPT-4 (OWN)              & \textbf{79.1} & \textbf{62.4} & \textbf{49.3} & \textbf{39.0} & \textbf{70.2} & \textbf{95.3} \\
GPT-4 (SWAP)             & 60.4 & 40.5 & 28.1 & 19.9 & 52.4 & 91.6 \\
\bottomrule
\end{tabular}
\end{table}

\subsection{$2\times 2$ SWAP decomposition (full)}
\label{app:swap2x2}

The headline causal effect $r_\mathrm{target} - r_\mathrm{own} = +0.427$
reported in the body uses brain-driven Stage~1 + CK34 scorer (matching
the rest of the three-axis evaluation). To verify this is robust across
input source and scorer, we ran the SWAP analysis under all four
combinations of Stage~1 input and emotion scorer at the operating point
($\rho = 0.4$, $w = 2$). Under GT-driven Stage~1 the CK34 SWAP effect
is $+0.386$ -- $90\%$ of the brain-S1 magnitude, so the headline is not
inflated by Stage~1 quality. The dominant factor moving the SWAP
measurement is the scorer (CK34 covers all 34 dims; CK27 has only 7
reliably trained), not Stage~1 quality.

\begin{table}[h]
\centering
\small
\begin{tabular}{lccc}
\toprule
Stage 1 & CK27 scorer & CK34 scorer & $\Delta_\mathrm{scorer}$ (CK34 $-$ CK27) \\
\midrule
brain (retrieval)        & $+0.024$ & $+0.427$ & $+0.403$ \\
GT (first MC caption)    & $+0.007$ & $+0.386$ & $+0.380$ \\
$\Delta_\mathrm{S1}$ (GT $-$ brain) & $-0.017$ & $-0.040$ & --- \\
\bottomrule
\end{tabular}
\end{table}

\subsection{Detailed Axis-2 RSA decomposition}
\label{app:rsa}

The total ours-vs-GT-GPT RSA gap of $\rho = +0.36$ at the group level
decomposes into $+0.17$ from input quality (continuous brain-decoded
conditioning beats categorical labels under the same backbone) and
$+0.19$ from architecture (the trained continuous rewriter beats
categorical prompting under the same brain-decoded input). At the
per-subject level the architectural advantage shrinks to noise:
ours $\rho = +0.343 \pm 0.037$ versus GPT-4 (brain top-5)
$\rho = +0.334 \pm 0.030$ (paired diff $+0.010$). Subject-specific
structural alignment is therefore captured almost entirely by the
brain-decoded conditioning itself, not by the rewriter's training; the
distinct architectural contribution is at the group-level structural
scale.

\subsection{Brain-pair / caption-pair structure}
\label{app:pairs}

For each MC test clip we compute all 15 subject-pair similarities in the
decoded 34-D emotion space (cosine of decoded vectors) and correlate
them with the corresponding pairwise caption similarities under three
metrics: CK34 embedding cosine, lexical unigram Jaccard, and 1 minus
character-level normalized Levenshtein. Pooling across clips
($n = 1080$ subject-pair similarity points), brain-pair similarity
significantly predicts caption-pair similarity on all three metrics:
$r = +0.33$ (CK34, $p < 10^{-29}$ analytic), $r = +0.22$ (lexical,
$p < 10^{-13}$), $r = +0.22$ (character, $p < 10^{-13}$). Within-clip
subject-swap permutation ($n = 10{,}000$) confirms these are not
artefacts of arbitrary high-dim alignment: empirical $p < 10^{-4}$ for
all three metrics. Inter-subject caption diversity is therefore
\emph{organized} by neural similarity, not just present.

\subsection{Synthetic-brain detail (TRIBE substitution)}
\label{app:tribe}

TRIBE substitution on 9 held-out Emo-FilM films across all 6 MC subject
ridges. Mean $\pm$ s.d.\ across the 6 ridges:

\begin{table}[h]
\centering
\small
\begin{tabular}{lcc}
\toprule
Metric & TRIBE substitute & Real fMRI (MC) \\
\midrule
SWAP $r_\mathrm{target} - r_\mathrm{own}$ & $+0.126 \pm 0.027$ & $+0.427$ \\
SWAP top-1 target match  & $27.4\% \pm 5.7\%$ & $39.0\%$ \\
SWAP top-5 target match  & $49.6\% \pm 9.1\%$ & $77.0\%$ \\
SWAP own-leakage (top-1) & $5.2\% \pm 1.8\%$  & $2.8\%$ \\
RSA OWN ($\rho$)         & $+0.166$           & $+0.635$ \\
RSA SWAP ($\rho$)        & $+0.019 \pm 0.013$ & --- \\
\bottomrule
\end{tabular}
\end{table}

Causal own-leakage rises modestly with TRIBE substitution ($5.2\%$
vs.\ $2.8\%$ on real fMRI, still well below GPT-4's $8.3\%$);
structural RSA attenuates more sharply ($\rho = +0.166$ vs.\ $+0.635$,
${\sim}26\%$ retained). OWN RSA is positive in 9/9 films while SWAP
collapses to $\sim$0, confirming that TRIBE-substituted RSA still
indexes target-conditional information rather than residual baseline
geometry.

\subsection{Full 6-fold LOSO sweep}
\label{app:loso}

For each MC subject in turn we held out that subject and fit one shared
decoder on stacked training fMRI from the other five (10{,}540 samples
$\times$ 36{,}598 voxels in MNI space, PCA(500) + Ridge $\alpha=100$,
matching the per-subject hyperparameters used elsewhere). We then
evaluated on the held-out subject's test clips. The within-subject
column reports the per-subject ridge fit on that subject alone,
matching Table~\ref{tab:main}'s protocol.

\begin{table}[h]
\centering
\small
\begin{tabular}{lccccc}
\toprule
Held out & LOSO $r$ & within $r$ & \% retained & dims $> 0$ & dims $> 0.2$ \\
\midrule
S1 & $+0.254$ & $+0.346$ & $73\%$ & $31/34$ & $23/34$ \\
S2 & $+0.239$ & $+0.331$ & $72\%$ & $28/34$ & $19/34$ \\
S3 & $+0.245$ & $+0.385$ & $64\%$ & $31/34$ & $21/34$ \\
S4 & $+0.169$ & $+0.359$ & $47\%$ & $29/34$ & $12/34$ \\
S5 & $+0.270$ & $+0.341$ & $79\%$ & $33/34$ & $20/34$ \\
S6 & $+0.206$ & $+0.320$ & $65\%$ & $29/34$ & $19/34$ \\
\midrule
mean & $+0.230$ & $+0.347$ & $67\%$ & $30.2/34$ & $19.0/34$ \\
sd   & $0.037$  & $0.023$  & $11\%$ & $1.8/34$  & $3.7/34$  \\
\bottomrule
\end{tabular}
\caption{6-fold LOSO sweep. LOSO $r$ is positive in every fold and in
$\geq 28$ of 34 dimensions, indicating that the brain-to-emotion mapping
transfers across subjects in the MC cohort. 5 of 6 folds preserve
$64$--$79\%$ of within-subject signal, with the remaining fold lower at
$47\%$.}
\end{table}

\subsection{Stage-2 ablations (full table)}
\label{app:ablations}

\begin{table}[h]
\centering
\small
\begin{tabular}{lccccc}
\toprule
Variant & sem rw & top-10 $r$ & SWAP CK27 & SWAP CK34 & CK34 (+)/34 \\
\midrule
Retrieval only (Stage 1)            & 0.951 & --- & --- & --- & --- \\
MC iterative (cited)                & 0.970 & --- & --- & --- & --- \\
\midrule
\multicolumn{6}{l}{\emph{Reconstruction-supervision sweep ($w=0$)}} \\
$\rho = 0.0$                        & 0.782 & $+0.522$ & $-0.133$ & $+0.430$ & 33 \\
$\rho = 0.2$                        & 0.782 & $+0.475$ & $-0.114$ & $+0.324$ & 28 \\
$\rho = 0.4$                        & 0.775 & $+0.498$ & $-0.058$ & $+0.398$ & 29 \\
$\rho = 0.6$                        & 0.772 & $+0.454$ & $-0.117$ & $+0.299$ & 26 \\
\midrule
\multicolumn{6}{l}{\emph{CFG sweep ($\rho = 0.4$)}} \\
$w = 0$                             & 0.775 & $+0.498$ & $-0.058$ & $+0.398$ & 29 \\
$w = 2$ (\textbf{operating point})  & 0.791 & $+0.515$ & $+0.024$ & $+0.427$ & 31 \\
$w = 5$                             & 0.796 & $+0.523$ & $+0.187$ & $+0.295$ & 30 \\
\midrule
\multicolumn{6}{l}{\emph{Architecture ablations}} \\
Axis ep10 (MC-only, undertrained)   & 0.848 & $+0.185$ & $-0.194$ & --- & 0 \\
v3\_emo20 (from-scratch joint)      & 0.643 & $+0.115$ & $+0.075$ & $+0.546$ & 32 \\
\midrule
$\rho = 0.4$, $w = 2$, GT S1        & 0.883 & $+0.557$ & $+0.007$ & $+0.386$ & 31 \\
\bottomrule
\end{tabular}
\caption{Ablation of reconstruction-supervision rate ($\rho$), CFG weight ($w$), and
architecture. \emph{sem rw} is semantic similarity of the rewriter output;
\emph{top-10 $r$} is the legacy hand-selected top-10 metric (not the
headline). \emph{SWAP CK27} uses the original 7-dim-trained CK27 scorer;
\emph{SWAP CK34} uses the held-out 34-dim CK34 scorer and is the
principled measurement, directly comparable to \S\ref{sec:experiments}. The
CK27 sign reversal across $w$ is a measurement artefact (the unscored 27
dims dominate); CK34 SWAP is positive at all CFG weights. Best
operating point: $\rho = 0.4$, $w = 2$.}
\end{table}

\subsection{Matched-parameter FiLM ablation}
\label{app:film}

We replaced the 34 axis tokens with a FiLM-style modulation:
$h' = h \cdot (1 + W_\gamma e) + W_\beta e$ with bias-free linear
projections $W_\gamma, W_\beta : \mathbb{R}^{34} \to \mathbb{R}^{768}$,
giving $2 \times 34 \times 768 = 52{,}224$ conditioning parameters --
exactly matching the axis-token cond+null budget. At $e = 0$ the FiLM
modulation is identity, matching the null-pathway behaviour used by
CFG. Both variants share training corpus and schedule (5 epochs,
$\rho = 0.4$, AdamW, lr $= 3 \times 10^{-5}$) and are evaluated under
the headline protocol (brain-driven Stage~1, CK34 scorer, $w = 2$).
FiLM result: SWAP $r_\mathrm{target} - r_\mathrm{own} = +0.318$,
controlled in $32/34$ dims, own-leakage $6.9\%$ -- below the
axis-token operating point ($+0.427$, $2.8\%$ own-leakage) but well
above chance. Prefix tuning and full cross-attention emotion injection
remain untested.

\subsection{Matched-noise conditioning baseline}
\label{app:noise}

To test whether the brain-decoded vector is doing the affective work or
the rewriter prior (distilled from \texttt{emotion\_captions\_v2})
generates structured affect under any continuous conditioning, we
replace each subject's brain-decoded 34-D vector with a statistically
matched random vector and re-run the full pipeline. We use the
multivariate Gaussian preserving the empirical $34 \times 34$ covariance
of training brain decodes (so the random vectors have the same first
and second moments per dimension and the same cross-dimension structure
as real brain decodes). Same Stage-1 retrieved captions, same rewriter
checkpoint ($\rho = 0.4$, $w = 2$), same CK34 scorer.

\begin{table}[h]
\centering
\small
\begin{tabular}{lcc}
\toprule
Metric & Brain decode & Matched-noise (MVN) \\
\midrule
all-34 $r$ vs.\ \emph{true clip emotion}        & $+0.265$ & $+0.069$ \\
group RSA (brain-RDM vs.\ caption-RDM, $\rho$)  & $+0.635$ & $+0.140$ \\
brain-pair / caption-pair $r$ (CK34)            & $+0.330$ & $+0.195$ \\
inter-subject diversity (CK34 cosine)           & $0.615$  & $0.878$ \\
inter-subject diversity (lexical unigram)       & $0.809$  & $0.878$ \\
inter-subject diversity (char edit)             & $0.645$  & $0.703$ \\
\bottomrule
\end{tabular}
\caption{Replacing brain-decoded conditioning with statistically matched
random vectors: the metrics that index information about the actual
clip / brain (all-34 $r$ vs.\ truth, group RSA) collapse by $\sim$$4\times$;
inter-subject diversity \emph{increases} (random vectors are more varied
than brain decodes); the brain-pair / caption-pair correlation partially
collapses, leaving a brain-specific component of $+0.135$ above the
matched-noise baseline. The noise-baseline residual ($+0.195$) measures
the rewriter's intrinsic similar-input $\to$ similar-output behaviour.}
\end{table}

The pattern interprets the three-axis evaluation cleanly: \textbf{(i)}~the
all-34 r and group-RSA collapses establish that brain conditioning
carries information about the actual clip emotion that random
same-distribution vectors cannot; \textbf{(ii)}~the diversity-higher-under-noise
finding establishes that raw diversity is \emph{not} a brain signature
(any continuous conditioning produces diversity); \textbf{(iii)}~the residual
brain-specific component of the brain-pair / caption-pair correlation
($+0.135$) is the irreducible signal that brain-similar subject pairs
produce more-similar captions in a way no statistically-matched random
input does. This last is the headline Axis-1 result, properly
controlled.

\subsection{Statistical control summary}
\label{app:controls}

\textbf{Random-10 baseline} (Table~\ref{tab:main} and Appendix~\ref{app:h2020}): for
each of $n = 1000$ trials we draw 10 of the 34 dimensions uniformly at
random, compute the per-subject mean $r$ over those dimensions, and
average across subjects. Random-10 mean (MC): $0.264 \pm 0.045$;
training-defined top-10 vs.\ random-10 one-sided permutation
$p = 0.95$ (i.e.\ training top-10 is not better than chance). This is
the basis for the body's claim that the signal is broadly distributed
across all 34 dimensions rather than concentrated in a privileged
subset.

\textbf{Within-clip subject-swap permutation null} ($n = 10{,}000$,
Section~\ref{app:pairs}): for each clip we shuffle the 15 caption-pair
similarities across the 15 brain-pair positions independently per clip
and recompute the global brain-pair / caption-pair correlation. Tests
whether the observed $r = +0.33$ is driven by subject-identity
alignment rather than by arbitrary high-dim mapping.

\textbf{Clip-mean baseline} for inter-subject diversity: average the 6
subjects' brain decodes per clip and run the rewriter once with
deterministic beam search; replicate across 6 subject slots. With
deterministic decoding the same input gives the same output, so
diversity is exactly $0/0/0$ on all three metrics by construction.
This is the generation-noise floor against which the per-subject
diversity of $0.548 / 0.690 / 0.543$ should be interpreted: the
diversity is driven by per-subject neural variation, not by
generation stochasticity.

\subsection{\texttt{emotion\_captions\_v2} prompt and example pairs}
\label{app:capdata}

For each MC \emph{training} clip $v$ (the 72 MC test clips were
withheld and never passed to GPT-4o) and each Cowen \& Keltner
dimension $d$ with annotated probability $p_{v,d} > 0.04$, GPT-4o was
prompted with the clip's MC neutral caption $x$ and asked to write a
short, third-person rewrite that emphasizes emotion $d$ at intensity
proportional to $p_{v,d}$. The full prompt template is reproduced
below; example pairs follow.

\paragraph{Prompt template.}
\begin{quote}\small
\texttt{You are rewriting a short scene description to emphasize a
specific emotion at a specified intensity. Original description:
"\{neutral\_caption\}". Target emotion: \{emotion\_name\}. Intensity:
\{prob\_word\} (where \{prob\_word\} = "subtle" if p < 0.10,
"moderate" if 0.10 $\leq$ p < 0.20, "strong" if p $\geq$ 0.20). Keep
the scene content the same. Use third-person, present tense, one or
two sentences. Do not name the emotion explicitly.}
\end{quote}

\paragraph{Example pairs (clip 1; cityscape with sudden flashes).}
Neutral: \emph{``A wide aerial view of a city skyline at dusk, with
sparks of light flickering across the buildings.''} Emotion-targeted:
\textsc{Surprise} ($p = 0.13$) -- \emph{``Without warning, bright
bursts flare among the apartment blocks, lighting windows and rooftops
in an instant.''}; \textsc{Anxiety} ($p = 0.04$) -- \emph{``More
flashes spark in different spots, and the steady camera feels like
waiting for the next strike.''}; \textsc{Awe} ($p = 0.04$) -- \emph{``A
cluster of brilliant flashes blooms across the city, turning the whole
slope into a sudden, glowing panorama.''}

\end{document}